\documentclass[runningheads]{llncs}

\usepackage{eccv}

\usepackage{eccvabbrv}

\usepackage{graphicx}
\usepackage{booktabs}

\usepackage[accsupp]{axessibility}  

\usepackage{array}
\usepackage{arydshln}
\usepackage{multicol}
\usepackage{multirow}

\usepackage{hyperref}

\usepackage{orcidlink}

\begin{document}

\setlength{\textfloatsep}{5pt}
\setlength{\floatsep}{5pt}
\setlength{\intextsep}{0pt}
\setlength{\abovecaptionskip}{5pt}
\setlength{\belowcaptionskip}{5pt}

\title{Beyond Viewpoint: Robust 3D Object Recognition under Arbitrary Views through Joint Multi-Part Representation} 

\titlerunning{Beyond Viewpoint}

\author{Linlong Fan\orcidlink{0000-0002-6350-7875}\and
Ye Huang\orcidlink{0000-0001-5668-5529}\thanks{Corresponding author} \and
Yanqi Ge\orcidlink{0009-0000-0086-4958}\and
Wen Li\orcidlink{0000-0002-5559-8594} \and
Lixin Duan\orcidlink{0000-0002-0723-4016}
}


\institute{ Shenzhen Institute for
 Advanced Study, \\ University of Electronic Science and Technology of China \\
\email{fanlinlong703@163.com}}

\maketitle

\begin{abstract}
Existing view-based methods excel at recognizing 3D objects from predefined viewpoints, but their exploration of recognition under arbitrary views is limited.
This is a challenging and realistic setting because each object has different viewpoint positions and quantities, and their poses are not aligned.
However, most view-based methods, which aggregate multiple view features to obtain a global feature representation, hard to address 3D object recognition under arbitrary views.
Due to the unaligned inputs from arbitrary views, it is challenging to robustly aggregate features, leading to performance degradation.
In this paper, we introduce a novel Part-aware Network (PANet), which is a part-based representation, to address these issues.
This part-based representation aims to localize and understand different parts of 3D objects, such as airplane wings and tails.
It has properties such as viewpoint invariance and rotation robustness, which give it an advantage in addressing the 3D object recognition problem under arbitrary views.
Our results on benchmark datasets clearly demonstrate that our proposed method outperforms existing view-based aggregation baselines for the task of 3D object recognition under arbitrary views, even surpassing most fixed viewpoint methods.

\keywords{3D object recognition \and weakly-supervised learning}
\end{abstract}

\section{Introduction}

\begin{figure}[t]
    \centering 
    \includegraphics[width=\linewidth]{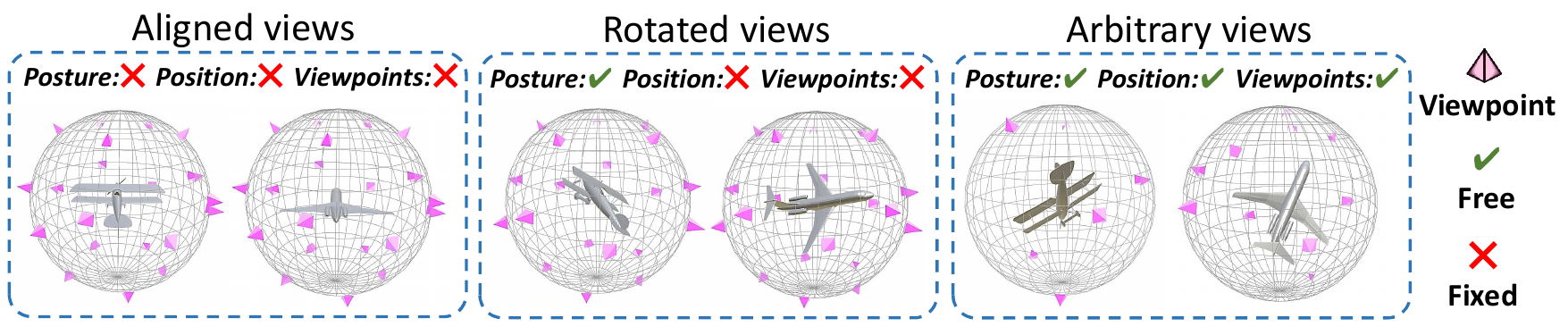}
    \caption{In practical scenarios, 3D objects are observed from arbitrary views\textbf{(right)}.
    The definition of arbitrary views is that each object is unaligned, and the viewpoint positions and quantities vary for each object.
    Previous works have focused on 3D object recognition in aligned views\textbf{(left)}, where the pose of each object is aligned, and the positions and quantities of viewpoints are predefined.
 Rotated views\textbf{(middle)} is an extension that introduces random rotation for each 3D object while keeping the viewpoint positions and quantities unchanged.
    Our work focuses on arbitrary views. }
    \label{fig:arbitrary_view}
\end{figure}

\begin{figure}[t]
    \centering
    \includegraphics[width=\linewidth]{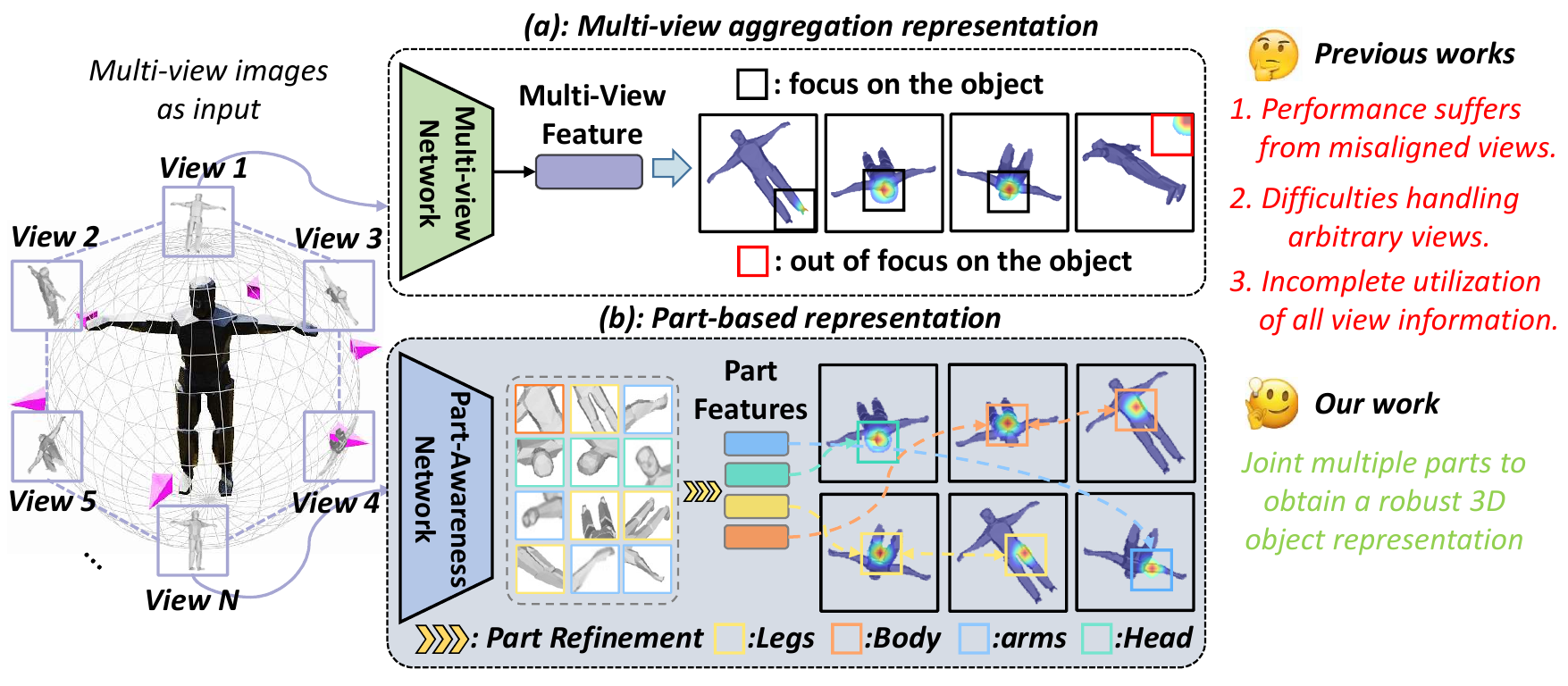}
    \caption{Comparison between multi-view aggregation representation \textbf{(a)} and part-based representation \textbf{(b)}.
    View aggregation methods integrate features from different views and use the aggregated features to represent 3D objects.
    However, since the views are not aligned, the aggregated features lack robustness, making it difficult to handle 3D recognition on arbitrary views.
    Additionally, view aggregation methods suffer from information loss, thus unable to fully utilize the features from each view.
    We newly propose a part-based representation that focuses on multi-view part awareness and combines multiple parts to robustly represent 3D objects.
     }
    \label{fig:PANet}
\end{figure}

Learning 3D with multi-view supervision has received widespread attention.
It covers various research domains such as NeRFs~\cite{nerf,nerfs_works}, BEV detection~\cite{bevformer,bevformerv2}, 3D generation~\cite{zero123}, 3D reasoning~\cite{3D_concept_learning, multi-CLIP}, etc., among which 3D object recognition is considered one of the most fundamental problems.
Thanks to mature and transferable 2D vision models~\cite{vit,stable_diffusion} and large-scale image datasets~\cite{imagenet}, rendering 3D objects into multiple 2D images and applying 2D networks to indirectly recognize 3D objects has become the state-of-the-art method~\cite{view-gcn++}.
Furthermore, compared with direct methods (\eg~, point-based~\cite{crosspoint,ulip}, voxel-based~\cite{voxnet}), view-based methods are closer to human perception of 3D objects in the real world.

Previous view-based methods~\cite{drcnn,hear,view-gcn,mvcnn-new, mvt} focus on addressing 3D object recognition tasks under fixed viewpoints, while recognition under arbitrary views still lacks extensive research.
As shown in Fig.~\ref{fig:arbitrary_view}, arbitrary views represent a more challenging setting where each object may have different viewpoint positions and quantities, and the poses of each object are not aligned.

However, most view-based methods~\cite{view-gcn, mvcnn-new, gvcnn, mvt, rn, drcnn, rotationnet, hear, 3d2seqviews, MHBN, MVLADN, dan, cvr, carnet, view-gcn++}, which aggregate multiple view features to obtain a global feature representation, hard to effectively address 3D object recognition under arbitrary views.
This is because the misaligned nature of arbitrary views complicates the representation and aggregation of view features, which leads to performance degradation.
For example, when applied to arbitrary views, the performance of CVR~\cite{cvr} drops by 6.31\% and 14.23\% in ModelNet~\cite{modelnet} and ScanObjectNN~\cite{scanobjectnn} datasets.
Even some multi-view aggregation methods~\cite{view-gcn,rotationnet} cannot be applied to arbitrary views because they require fixed viewpoint position information.
Furthermore, view aggregation methods~\cite{mvcnn-new,gvcnn} typically jointly optimize feature extraction and aggregation modules to maximize performance, but this may lead the extraction module to overly focus on certain views while ignoring information from others, resulting in information loss.

To address these challenges, we propose a novel part-based representation, as shown in Fig.~\ref{fig:PANet}.
This idea is inspired by cognitive science and biology~\cite{cognitive}, which aims to understand different part features or specific regions in a multi-view sequence to better represent 3D objects.
For example, a person can be composed of multiple parts with unique attributes, such as the head, arms, body, legs, and so on. %
By locating and integrating the features of these parts, we can more accurately represent a person.
Compared with the multi-view aggregation features, part features offer several advantages.
Firstly, an individual view contains abundant part information. Even if only a few views are utilized, we can still extract stable part features, thereby adapting to changes in the number of viewpoints in arbitrary view settings.
Secondly, the part features are more stable than the multi-view aggregation features~\cite{sift}, which can better adapt to changes in object rotation and viewpoint positions.
Therefore, the part-based representation method is robust and can better handle the challenge of 3D object recognition under arbitrary views.
We also provide comprehensive experimental results in Sec.~\ref{sec:experiments} to demonstrate the effectiveness of part-based representation.
When switching to arbitrary views, we do not observe significant performance degradation in our method, and it even outperforms most fixed viewpoint methods.

Specifically, this paper proposes a novel part-aware network (PANet).
Firstly, it employs weakly supervised learning to locate and recognize parts in each view, obtaining local view part features.
Then, local view parts are input into the adaptive part refinement~(APR) module.
This module adaptively perceives the correlation of local view parts and further integrates these view parts to obtain global parts, which aims to adapt to the viewpoint changes of parts across different views.
Finally, we combine these global parts to represent a 3D object.
It is worth noting that we also introduce a regularization loss in the part localization module to ensure that part features can better focus on the object.

The primary contributions of this work can be summarized as follows: 

$-$ Unlike existing view aggregation methods, we propose a novel part-based representation. We consider a 3D object as a collection of part features rather than an abstract multi-view aggregated feature. This part-based representation is not limited by the position and quantity of viewpoints or affected by object poses, making it more flexibly applicable to various viewpoint scenarios, especially in arbitrary views.

$-$ Further, we newly develop a Part-Aware Network~(PANet), for extracting parts in multiple views. 
The network employs a weakly supervised method to localize parts on each view, enabling it to obtain local view parts. 
Then, we further refines these view parts through the proposed Adaptive Part Refinement (APR) module to obtain global parts.
These global parts can better adapt to differences in parts across viewpoints.
Finally, integrating these global parts forms a comprehensive description of a 3D object.

$-$ We conduct experiments on various datasets, including ScanObjectNN, ModelNet, and RGBD. Our method outperforms existing methods in both fixed views and arbitrary views. In particular, the performance in arbitrary views exceeds most existing fixed viewpoint methods based on fixed views, achieving robust recognition of 3D objects.

\section{Relative Work}

{\noindent\textbf{3D object Recognition methods: }}
3D object recognition using deep neural networks can be broadly divided into three main categories: point-based methods~\cite{pointnet,pointnet++,3D_survey}, voxel-based methods~\cite{voxnet,lp-3dcnn,octnet}, and view-based methods.
View-based methods render 3D objects into multiple views and utilize multiple views to jointly represent 3D objects.
Benefiting from mature and transferable 2D vision models and large-scale image datasets, they can better handle 3D object recognition tasks.

MVCNN~\cite{mvcnn} is a pioneering multi-view network for 3D object classification. 
It utilizes a view pooling strategy to integrate features from different views. 
Yang \etal~\cite{rn} proposes a relational network to effectively connect corresponding regions from different viewpoints, thus enhancing the information of each view.
3D2Seq2views~\cite{3d2seqviews} method treats multiple views as a sequence and employs a convolutional neural network with novel hierarchical attention aggregation to aggregate sequential views. 
DRCNN~\cite{drcnn} utilizes the dynamic routing algorithm of capsule networks to construct a new Dynamic Routing Layer~(DRL), which integrates features from each view.
View-GCN~\cite{view-gcn} employs multiple views as nodes in a view graph, leveraging a graph convolutional network (GCN) architecture to hierarchically aggregate information and learn robust 3D shape descriptors.
These methods assume that each object has a predefined set of viewpoints and are trained and tested on these viewpoints.
This assumption may limit model performance when dealing with 3D object recognition in arbitrary views.
As far as we know, there are few studies~\cite{mvtn, cvr, dcp} that have made significant progress beyond fixed viewpoint settings.
The MVTN~\cite{mvtn} implements viewpoint regression for 3D object recognition based on the latest research progress of differentiable rendering.
The motivation of CVR~\cite{cvr} is to restore the intrinsic alignment of arbitrary views, transform it into aligned representations and then perform aggregation.
In our work, we newly propose part-based representation.
Part features exhibit viewpoint invariance and rotation robustness, making them more suitable for handling changes in viewpoint positions and quantities in arbitrary views.
Moreover, the key idea of part-based representation is not to rely on multi-view aggregation, thus avoiding information loss.

{\noindent\textbf{Discriminative Part Localization: }}
The perception and localization of object part-level features are usually challenging. 
Many existing approaches~\cite{fine_3d,huang2016part,zhang2016spda,zhang2014part} suggest using bounding boxes and additional annotations to accurately locate discriminative regions of objects.
However, this method often requires expert knowledge and extensive annotation efforts, rendering it impractical for real-world applications. 
Fortunately, more and more research~\cite{zhou2016learning,racnn,macnn,wsdan,counterfactual_attention} has shown that important object regions can also be localized using weakly supervised methods.
These methods rely solely on image-level supervision to identify part features.
Early works usually locate class-specific discriminative regions via CAM~\cite{zhou2016learning}.
Fu~\etal proposes Recursive Attention CNN (RACNN)~\cite{racnn} to iteratively focus on predicting the locations of attention regions and extract discriminative features.
Zheng~\etal proposes Multi-Attention CNN (MACNN)~\cite{macnn}, which proposes channel grouping loss to generate multiple parts by clustering.
WSDAN~\cite{wsdan} proposes weakly supervised attention learning to generate the distribution of parts.
Meng~\etal~\cite{pdm} introduces an end-to-end part discovery model (PDM) that aims to uncover robust and diverse object parts within a unified framework.
Our PANet focuses on perceiving multiple discriminative regions in each view, extracting sufficient local view parts, and proposing the Adaptive Part Refinement~(APR) module to further refine them to obtain global parts.
These global parts can adapt to changes in viewpoint for parts across different views, making them more suitable for 3D objects.

\section{Proposed Method}
\begin{figure*}[t]
    \centering       
    \includegraphics[width=\linewidth]{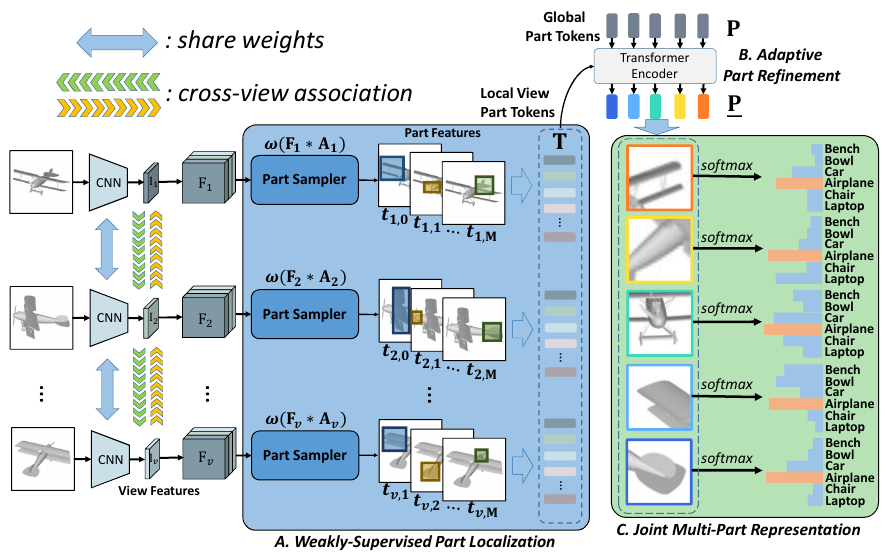}
    \caption{Overview of our proposed Part-Aware Network (PANet). 
    The network takes multi-view images as input, encodes them by CNN, obtains view features $\textbf{I}$, and uses the cross-view association (CVA) module to enhance the view features.
    Firstly, employ weakly-supervised methods to perceive part regions and generate view part sequences $\textbf{T}$.
    Then, given some learnable part tokens $\textbf{P}$, the sequences $\textbf{T}$ and $\textbf{P}$ are concatenated and input to the adaptive part refinement (APR) module.
    The module aims to refine the sequence $\textbf{T}$ into a more compact representation, resulting in the global part features.
    Finally, multiple global parts $\underline{\textbf{P}}$ together serve as a representation of a 3D object.}
    \label{fig:framework}
\end{figure*}

In this work, we newly propose a Part-Aware Network~(PANet) that aims to solve the challenge of 3D object recognition (especially under arbitrary views).
The overall architecture of the proposed PANet is shown in Fig.~\ref{fig:framework}, consisting of three modules: weakly supervised part localization, adaptive part refinement, and joint part representation.
We first introduce a weakly supervised method to perform part localization and feature extraction on each view (Sec.~\ref{sec:part-aware}).
Secondly, to adapt to the variations of part features across views, we propose the Adaptive Part Refinement (APR) module, which refines redundant part features from multiple views into more compact representations (Sec.~\ref{sec:apr}).
Finally, we integrate these refined features to represent a 3D object (Sec.~\ref{sec:joint}).

\subsection{Weakly-Supervised Part Localization}
\label{sec:part-aware}
It is worth noting that we do not rely on supervised positional annotations~\cite{zhang2014part}, such as bounding boxes or keypoints, for part localization.
Instead, we employ a more cost-efficient weakly-supervised learning approach~\cite{wsdan,macnn} that utilizes image-level annotations to identify discriminative regions, which requires no expertise and no extensive annotation effort.

We generate a multi-view sequence $\mathcal{X} = \{\textbf{X}_{1}, \textbf{X}_{2},\ldots,\textbf{X}_{v}\}$ by rendering a 3D object from different viewpoints. 
Here, $v$ denotes the number of views, $\textbf{X}_{i}$ represents the $i$-th view, where $i \in \left[1,v\right]$.
Then, each view is encoded by an image classification network \eg~, ResNet~\cite{resnet} to obtain view features $\textbf{I}={\{\textbf{I}_1, \textbf{I}_2,\ldots,\textbf{I}_{v}\}}$, where $\textbf{I}_{i} \in \mathbb{R}^{{\rm H} \times {\rm W} \times {\rm C}}
~\footnote{H, W, and C represent the height, width, and
number of channels of the feature layer, respectively}, i \in \left[1,v\right]$ 
is corresponding to the feature map of the $i$-th view $\textbf{X}_i$.
To align the features from multiple views into the same feature space, we share the weights among each of the convolutional neural network.

\noindent\textbf{Enhance Multi-view Features: }
However, the associations between view features $\textbf{I}$ are not explicitly modeled.
Therefore, we develop a cross-view association~(CVA) module to obtain enhanced view features $\textbf{F}_i$, enabling each view to incorporate information from other views.

For each view feature $\textbf{I}_i$, the cross-view association~(CVA) module calculates the matching score between it and the overall view feature $\textbf{I}$.
Here, the dot product function is selected as the matching function.
Then, $\textbf{I}_i$ is enhanced based on the correlation scores with other views.
The enhanced view feature $\textbf{F}_{i} \in \mathbb{R}^{{\rm H} \times {\rm W} \times {\rm C}}$ can be expressed in detail as follows:
\begin{equation}
    \textbf{F}_{i} ={softmax\left({\omega(\textbf{I}_{i})^{T} \cdot \omega(\textbf{I})}\right)\textbf{I}},
\label{eq:F}
\end{equation}
where $\omega(\textbf{I}_{i}) \in \mathbb{R}^{{\rm C} }$ represents the use of average pooling operations.

\noindent\textbf{Part Sampler: } 
After passing through the cross-view association module, the enhanced view features $\textbf{F}$ comprehensively incorporate information from other views.
Next, we sample parts from the enhanced view features $\textbf{F}$.

Inspired by~\cite{counterfactual_attention,wsdan}, we implement the attention model $\psi$ to learn the spatial distribution of each part within each view.
For view $i$, the distribution of view parts is represented by the attention map $\textbf{A}_{i} \in \mathbb{R}^{{\rm H} \times {\rm W} \times {\rm M}}$, which can be computed by:
\begin{equation}
    \textbf{A}_{i} = \psi(\textbf{F}_{i}) = \left\{ \textbf{A}_{i,1},\textbf{A}_{i,2},\ldots,\textbf{A}_{i,\rm M} \right\},
\label{eq:attention}
\end{equation}
where $\textbf{A}_{i,j} \in ~\mathbb{R}^{{\rm H} \times {\rm W}}, i \in \left[1,v\right], j \in \left[1,\rm M\right]$ represents the $j$-th part region or visual pattern of the object's $i$-th views, such as a car wheel or an airplane wing.
$\rm M$ denotes the number of attention maps, each view extracts the same number of regions of interest. 

Next, we employ the attention map $\textbf{A}_{i,j}$ to softly weight the enhanced feature maps $\textbf{F}_i$, followed by the global average pooling operation $\omega$ to obtain the parts on view $i$ (See Eq.~\ref{eq:tij}).
\begin{equation}
\boldsymbol{t}_{i,j} = \omega\left( {\textbf{F}_{i}\otimes\textbf{A}_{i,j}} \right),
\label{eq:tij}
\end{equation}
where $\boldsymbol{t}_{i,j} \in \mathbb{R}^{{\rm C}}$, $\otimes$ denotes element-wise multiplication between two tensors.

Finally, we concatenate all the view part features together, resulting in a 3D multi-view part sequence $\textbf{T} \in~\mathbb{R}^{\rm N \times C}$ (see Eq.~\ref{eq:T}).
\begin{equation}
    \begin{aligned}
    \textbf{T} = \lbrack
        \boldsymbol{t}_{0,0}, \boldsymbol{t}_{0,1}, \ldots, \boldsymbol{t}_{0,{\rm M}}, \boldsymbol{t}_{1,0}, \boldsymbol{t}_{1,1}, \ldots, \boldsymbol{t}_{1,{\rm M}}, ..., \boldsymbol{t}_{v,0}, \boldsymbol{t}_{v,1}, \ldots, \boldsymbol{t}_{v,{\rm M}} \rbrack,
    \end{aligned}
\label{eq:T}
\end{equation}
where $\rm N$ represents the total number of parts~$\left(v \times {\rm M} \right)$ in the multi-view.

\subsection{Adaptive Part Refinement}
\label{sec:apr}

In Sec.~\ref{sec:part-aware}, we have obtained a multi-view part sequence $\textbf{T}$.
However, there are differences in parts across different views, where each element $t_{ij}$ in $\textbf{T}$ can only represent local view features.
It is challenging to generalize the overall features of a 3D object relying solely on local view parts.
Therefore, we newly propose an Adaptive Part Refinement (APR) module, which aims to explicitly model the correlations between parts across views, thus enhancing the robustness of part features to viewpoint variations.
This module adaptively integrates part information from multiple views and refines it to obtain global part features that are more suitable for 3D objects.

Formally, the APR (Adaptive Part Refinement) module contains a transformer block. Given some global part tokens~\cite{maskformer,vit,detr} $\textbf{P} \in \mathbb{R}^{\rm L \times C}$, where $\rm L$ represents the number of parts.
We concatenate $\textbf{P}$ and $\textbf{T}$ as input to the APR module.
After passing through the APR module, local parts $\boldsymbol{t}_{i,j}$ can learn cross-view correlations and obtain global parts $\textbf{\underline{P}} \in ~\mathbb{R}^{\rm L \times C}$ suitable for 3D objects.
We also observe the diversity of global part features $\textbf{\underline{P}}$~(See Fig.~\ref{fig:vis_part}).

\subsection{Joint Multi-Part Representation} 
\label{sec:joint}
Relying solely on individual parts may not be sufficient to represent a 3D object.
To overcome this limitation, we jointly describe the 3D object using multiple parts.
We make predictions for each part and then average them to obtain the prediction for the 3D object.
Each 3D object can be represented as Eq.~\ref{eq:Parts}:
%
\begin{equation}
    \textbf{\underline{P}} = \left[ \underline{\boldsymbol{p}}_1, \underline{\boldsymbol{p}}_2, \ldots , \underline{\boldsymbol{p}}_{\rm L}\right], ~~~\rm where~\textbf{\underline{P}} \in~\mathbb{R}^{\rm L \times C}
\label{eq:Parts}
\end{equation}

The proposed PANet is optimized through supervised learning, including instance classification loss $\mathcal{L} _{ce}$ and part-aware loss $\mathcal{L}_{awe}$.
The loss function for multi-view $\mathcal{X}$ is defined as follows:
\begin{equation}
\begin{split}
    &p^i = softmax\left(\underline{\boldsymbol{p}}_i  \boldsymbol{W}_p + \boldsymbol{b}_p \right),~~~{\rm where}~\underline{\boldsymbol{p}}_i \in \mathbb{R}^{\rm 1 \times C},~i \in \left[1,\rm L\right]\\
    &q^j = softmax\Big(\boldsymbol{t}_j \boldsymbol{W}_q + \boldsymbol{b}_q\Big), ~~~{\rm where} ~\boldsymbol{t}_j  \in \mathbb{R}^{\rm 1 \times (M \times C)},~j \in \left[1,v\right],\\  
    &\mathcal{L} = ~\mathcal{L}_{ce} + \mathcal{L}_{awe} 
      = -\sum_{k = 1}^{K}{y_k} {log\hat{p}_k}+\gamma \left(-\frac{1}{v} \sum_{i = 1}^{v}\sum_{k = 1}^{K} {y_k}{logq^j_k} \right),
\end{split}
\label{eq:loss}
\end{equation}
where $\boldsymbol{W}_p$, $\boldsymbol{W}_q$, $\boldsymbol{b}_p$, and $\boldsymbol{b}_q$ are learnable parameters.
$p^i$ represents the predicted probability of the $i$-th global part, $\hat{p}_k$ represents the average of the predicted probabilities for $\rm L$ global parts, and $q^j$ represents the predicted probability for the $j$-th view part.
$y_k$ is the ground truth label of class $k$. 
$\mathcal{L}_{ce}$ represents the cross-entropy loss of the final parts.
$\mathcal{L}_{awe}$ represents the classification loss of $\rm M$ part features in each view, which is used to constrain the generation of parts.
Furthermore, we apply techniques such as flipping and random erasure to augment the image for more robust part results.

\section{Experiments}
\label{sec:experiments}
\subsection{Datasets and Experimental Setup}
We conduct an extensive evaluation of PANet on several benchmark datasets, namely ScanObjectNN~\cite{scanobjectnn}, ModelNet~\cite{modelnet}, and RGBD~\cite{rgbd}.
In order to facilitate fair comparisons, we strictly adhere to the methodologies proposed in prior works~\cite{mvcnn,rotationnet,cvr} for preparing the training and testing data on the different datasets.
Specifically, we adopt the training scheme introduced by MVCNN~\cite{mvcnn} and adopt the same viewpoint generation and rendering method as RotationNet~\cite{rotationnet} to handle the multi-view images from the ModelNet and ScanObjectNN datasets.
Experiment on rotated datasets also follows CVR~\cite{cvr} setting.
For arbitrary views, the number of views for each 3D object ranges randomly from 10 to 20.
Throughout the training process, we utilize the AdamW~\cite{adamw} optimizer with a learning rate of 1e-5.
We select ResNet50 as our backbone network and utilize label smoothing regularization~\cite{label_smoothing} to enhance the generalization capability of our model.
$\gamma$ is the weighting factor for the part-aware loss as in Eq.~\ref{eq:loss}, which we set $\gamma = 1$ for the experiments. 
Please refer to the Appendix for implementation details of our method.
\subsection{Comparison with State-of-the-Art Method}
%
%
\begin{table*}[tp]
        \caption{Classification on ScanObjectNN. `Aligned' means that every 3D object is in the same pose.  `Rotated' means that each 3D object is misaligned. `-' indicates that the specific result is not reported.}
        \scriptsize
        \centering
        \setlength{\tabcolsep}{1.5mm}{
        \begin{tabular}{lccccc}
        \toprule
                \multirow{2}{*}{Method} & \multirow{2}{*}{Type} & \multicolumn{2}{c}{Aligned}  & \multicolumn{2}{c}{Rotated}      
                \\ 
                \cmidrule{3-6}
                 & & {Per Class Acc.} & {Per Ins. Acc.} & {Per Class Acc.} & {Per Ins. Acc.}    \\ \midrule
                PointNet++~\cite{pointnet++}     &   \multirow{2}{*}{Point}   &  82.10\% & 84.30\%     &  -  &   -   \\ 
                DGCNN~\cite{dgcnn}     &    &   84.00\%  & 86.20\%     &  -  &   -   \\ 
                 \midrule
                MVCNN~\cite{mvcnn}     &   \multirow{6}{*}{View}   &  85.71\%  &   87.82\%   &  78.21\%  &   80.62\%   \\ 
                GVCNN~\cite{gvcnn}     &  \multirow{6}{*}{(fixed)}    &  86.64\%  &   88.68\%   &  82.86\%  &   83.70\%   \\ 
          RotationNet~\cite{rotationnet}   &    &  84.88\%  &   86.90\%    &  74.68\%  &      76.16\%   \\ 
                View-GCN~\cite{view-gcn}   &      &  88.67\%  &   90.39\%   &  81.99\%  &   83.50\%   \\ 
                CVR~\cite{cvr}             &   &  88.39\%  &   90.74\%   &  84.70\%  &   85.59\%   \\
                View-GCN++~\cite{view-gcn++}   &    &  89.10\%  &   91.30\%   &  85.80\%  &   86.80\%   \\
                \textbf{Ours}    &      &    \textbf{93.21\%}    &     \textbf{93.99\%}    &  \textbf{86.15\%}  &   \textbf{87.82\%} \\
                \midrule
                CVR~\cite{cvr}       & \multirow{2}{*}{(arbitrary)}  &  -  &   -   &  68.07\%  &   71.36\%   \\ 
                \textbf{Ours}    &       &    91.77\%    &     93.48\%     &  84.08\%  &   86.45\% \\
        \bottomrule
        \end{tabular}}
\label{tab:scanobjectnn}
\end{table*}

\begin{table}[tb]
\parbox[t]{.48\linewidth}{
\centering
\caption{Classification on Aligned ModelNet40. `-' indicates that the specific result is not reported, `fixed' denotes sampling from a fixed viewpoint. `free' means that the arbitrary number of viewpoints.}
\label{tab:align_modelnet}
\setlength{\tabcolsep}{-1pt}
\scriptsize
\begin{tabular}{lccc}
            \toprule
            & \multicolumn{2}{c}{Aligned ModelNet40}  
            \\ \cmidrule{1-4} 
            Method & Type & {Class Acc.} & {Ins. Acc.}
            \\ \midrule
            3DShapeNets~\cite{shapenet}    &   \multirow{3}{*}{Voxel}   &   77.3\%  &   -  \\ 
            VoxNet~\cite{voxnet}    &      &  83.0\%  &   -  \\ 
            VRNEnsemble~\cite{VRNEnsemble}    &     &  91.4\%  &   93.8\%  \\ 
            \midrule
            PointNet++~\cite{pointnet++}    &   \multirow{5}{*}{Point}   &  -  &   91.9\%  \\ 
            Kd-Networks~\cite{kd-networks}    &      &  88.5\%  &   91.8\%  \\ 
            RS-CNN~\cite{rscnn}    &     &  -  &   93.6\%  \\ 
            KP-Convs~\cite{kpconv}    &      &  -  &   92.9\%  \\ 
            MVTN~\cite{mvtn}    &     &  92.0\%  &   93.8\%  \\ 
            \midrule
            MVCNN-new~\cite{mvcnn}    &      &  92.40\%  &   95.00\%  \\ 
            GVCNN~\cite{gvcnn}          &    &  90.70\%  &   93.10\%   \\
            HEAR~\cite{hear}                      &   &  95.20\%  &   96.70\%   \\ 
            RotationNet~\cite{rotationnet}   &     &    -      &   97.37\%   \\ 
            View-GCN~\cite{view-gcn}   &   &  96.50\%  &   97.60\%   \\ 
            MVT~\cite{mvt}                     &     &  -  &   97.50\%   \\ 
             HMVCM~\cite{hmvcm}                      &   & -   &   94.57\%   \\ 
            MVD~\cite{mvd}                      &  View &  92.61\%  &   94.41\%   \\ 
            DAN~\cite{dan}   &  &  -  &   93.50\%  \\
            DRCNN~\cite{drcnn}                      &   &  94.86\%  &   96.84\%   \\ 
             View-GCN++~\cite{view-gcn++}                      &     &  96.50\%  &   97.60\%     \\
            CVR~\cite{cvr}            &  &  95.77\%  &   97.16\%   \\ 
            HGNN+~\cite{hgnn+}                 &   & -  &   96.92\%   \\ 
            
            \textbf{Ours}    &    &    \textbf{96.60\%}    &     \textbf{97.73\%}    \\
            \midrule
            \textbf{Ours}    &  (free)  &  96.37\%  &   97.53\%     \\
        \bottomrule
        \end{tabular}
}
\hfill
\parbox[t]{.48\linewidth}{
\centering
\begin{minipage}[t]{1\linewidth}
\centering
\scriptsize
\caption{ The accuracy of 3D object classification on Rotated ModelNet40.}
\label{tab:rotation_modelnet}
\setlength{\tabcolsep}{-1pt}
\begin{tabular}{lccc}
            \toprule
            & \multicolumn{2}{c}{Rotated ModelNet40}  
            \\ \cmidrule{1-4} 
            Method & \#View & {Class Acc.} & {Ins. Acc.}\\
            \midrule
            MVCNN~\cite{mvcnn}    &  \multirow{8}{*}{fixed} &  87.95\%  &   88.17\% \\ 
            GVCNN~\cite{gvcnn}         &    &  89.69\%  &   88.10\%   \\ 
            EMV~\cite{equivariant}      &  &      88.80\%  &  90.00\%  \\
            RotationNet~\cite{rotationnet}   &     &  84.74\%  &                85.29\%    \\ 
            View-GCN~\cite{view-gcn}   &   &  85.90\%  &   88.25\%  \\ 
            CVR~\cite{cvr}                        &   &  \textbf{91.12\%}  &   92.22\%      \\
            View-GCN++~\cite{view-gcn++}   &   &  90.70\%  &   92.80\%  \\
              
            \textbf{Ours}    &  & 90.35\%    &     \textbf{92.87\%}   \\
            \midrule
            CVR~\cite{cvr}                        &  \multirow{2}{*}{arbitrary}  &  84.01\%   &   86.91\%      \\
             \textbf{Ours}    &    &  90.69\%  &   92.23\%    \\
        \bottomrule
        \end{tabular}
\end{minipage}%
 
\begin{minipage}[t]{1\linewidth}
\centering
\captionof{table}{The accuracy of object classification on RGBD (in\%).}
\label{tab:rgbd}
\scriptsize
\setlength{\tabcolsep}{7pt}
\begin{tabular}{l c c}
    \toprule
Method       & $\#$View                   & Ins. Acc. \\
    \midrule
MDSICNN~\cite{MDSICNN}       & $\geq$  120 & 89.6\%        \\
CFK\cite{CFK}       & $\geq$  120  & 86.8\%      \\
MMDCNN~\cite{NMDCNN}     &  $\geq$  120   & 86.8\%        \\
RotationNet~\cite{rotationnet}  & 12                  & 89.3\%     \\
View-GCN~\cite{view-gcn}    &            12           & 94.3\%         \\
View-GCN++~\cite{view-gcn++}    &            12           & 92.5\%         \\
    
\textbf{Ours}    &            12           & \textbf{96.5\%}         \\
    \midrule
MVCNN~\cite{mvcnn}    & \multirow{4}{*}{4-12}         &   89.0\%       \\
GVCNN~\cite{gvcnn}     &                     & 89.8\%            \\
CVR~\cite{cvr}     &                      &  91.8\%         \\
\textbf{Ours}         &                      &  \textbf{95.9\%}         \\
    \bottomrule
\end{tabular}
\end{minipage}
}
\end{table}

{\noindent\textbf{The ScanObjectNN Datasets.} 
As a recently introduced real-world 3D object classification dataset, ScanObjectNN~\cite{scanobjectnn} provides scanned indoor scene data. 
The dataset comprises approximately 15,000 objects, spanning 15 categories, and includes 2,902 unique object instances. 
Compared with CAD handcrafted datasets, ScanObjectNN offers more realistic challenges as it reflects the noise, complexity, and incompleteness present in real-world environments, incorporating the appearance and geometric shapes of actual objects. 
Consequently, it is better for assessing the robustness of the model.
We train PANet under different viewpoint settings and compare with point-based methods~\cite{pointnet++,dgcnn} and view-based methods~\cite{mvcnn,gvcnn,rotationnet,view-gcn,cvr,view-gcn++}, as shown in Tab.~\ref{tab:scanobjectnn}.
View-based methods primarily rely on view aggregation, and it is difficult to robustly aggregate features from structurally misaligned views, especially for arbitrary views.
When switching to arbitrary views, the performance of these methods drops sharply.
Our PANet is observed to significantly outperform previous approaches in both class accuracy and instance accuracy, whether in aligned views, rotated views, or arbitrary views.
In aligned views, compared with the current state-of-the-art methods \eg~, View-GCN++~\cite{view-gcn++} and CVR~\cite{cvr}, PANet improves instance accuracy and class accuracy by 2.69\% and 4.11\%.
PANet also achieves significant improvements of 1.02\% and 0.35\% in rotated views.
Surprisingly, even under arbitrary views, our performance surpasses that of most fixed viewpoint methods, demonstrating that PANet can robustly recognize 3D objects unconstrained by the scene.
Please refer to the Appendix for more detailed comparisons.
}

{\noindent\textbf{The ModelNet Datasets.} 
ModelNet~\cite{modelnet} is a dataset of hand-drawn point clouds generated from CAD models.
It comprises 12,311 3D shapes, covering 40 categories, with 9,483 models allocated for training and 2,468 for testing.
The experimental results on the ModelNet dataset are presented in Tab.~\ref{tab:align_modelnet}.
We first compare with point-based and voxel-based methods, including PointNet++~\cite{pointnet++}, Kd-Networks~\cite{kd-networks}, RS-CNN~\cite{rscnn}, KP-Convs~\cite{kpconv}, MVTN~\cite{mvtn}, 3DShapeNets~\cite{shapenet}, VoxNet~\cite{voxnet}, and VRNEnsemble~\cite{VRNEnsemble}.
Our PANet outperforms them by more than 3.93\% and 4.6\% in instance accuracy and class accuracy and is also 3.73\% and 4.37\% higher in free viewpoints.
We also compared with view-based approaches, including MVCNN-new~\cite{mvcnn}, GVCNN~\cite{gvcnn}, HEAR~\cite{hear}, RotationNet~\cite{rotationnet}, View-GCN~\cite{view-gcn}, MVT~\cite{mvt}, HMVCM~\cite{hmvcm}, MVD~\cite{mvd}, DAN~\cite{dan}, DRCNN~\cite{drcnn}, EMV~\cite{equivariant}, View-GCN++~\cite{view-gcn++}, CVR~\cite{cvr}, and HGNN+~\cite{hgnn+}.
PANet consistently outperforms these methods in fixed views.
Its performance at free viewpoints surpasses that of most existing approaches.
We next report the robustness of PANet in rotated views, as shown in Tab.~\ref{tab:rotation_modelnet}.
Under the rotated views, PANet also achieves the highest score in instance accuracy.
}
{\noindent\textbf{The RGBD Datasets.} 
The RGBD~\cite{rgbd} dataset comprises 51 categories of domestic objects captured in the real world, encompassing RGBD views from multiple viewpoints. 
To ensure a fair comparison, we adopt the identical experimental setup as RotationNet~\cite{rotationnet} and View-GCN~\cite{view-gcn}. 
Specifically, we uniformly sample 12 RGB images from the same circle for each 3D object as input and report the average results using ten-fold cross-validation.
We keep the same settings as CVR~\cite{cvr} in arbitrary views, sampling images from 4 to 12 viewpoints as input to the network.
The experimental results are summarized in Tab.~\ref{tab:rgbd}. 
The performance of PANet in arbitrary views is impressive.
Compared with work in arbitrary views \eg~, CVR~\cite{cvr}, PANet significantly improves by 4.1\%.
Additionally, compared with the work in fixed views, such as view-GCN~\cite{view-gcn}, PANet has also improved by 1.6\%.
This further validates the superiority of our method.}

\begin{figure*}[t]
\begin{minipage}{.48\linewidth}
\centering
\includegraphics[width=5.5cm]{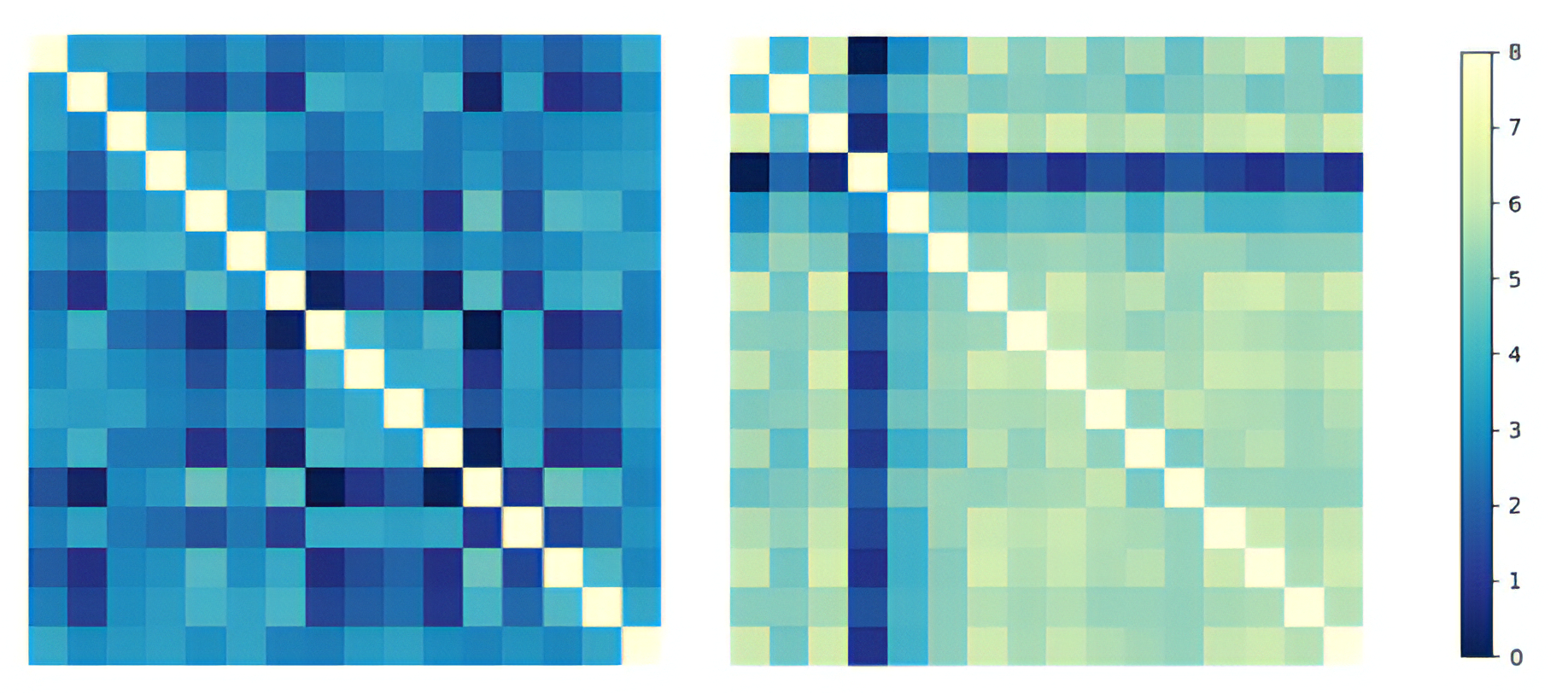}
\caption{Visualization of correlations between part features $\underline{\textbf{P}}$. The left figure illustrates the correlation between $\underline{\textbf{P}}$ after applying $\mathcal{L}_{awe}$.}
\label{fig:vis_part}
\end{minipage}
\noindent
\hfill
\begin{minipage}{.48\linewidth}
\setlength{\abovecaptionskip}{-5pt}
\setlength{\belowcaptionskip}{15pt}
\centering
\scriptsize
\setlength{\tabcolsep}{8pt}
\captionof{table}{PANet component ablations. The first line serves as the baseline. $\mathcal{L}_{awe}$ refers to the weakly supervised part-aware loss function. CVA stands for Cross-View Association module.}
\begin{tabular}{c c c c}
            \toprule
             CVA &   $\mathcal{L}_{awe}$       &  Class Acc.  & Ins. Acc.  \\
            \midrule
            &  & {83.76}\%         & {84.56}\%       \\ 
            \checkmark &  & {84.99}\%         & {85.25}\%       \\ 
            & \checkmark  & {85.45}\%         & {86.96}\%       \\ 
            \bottomrule
\end{tabular}
\label{tab:component_ablation}
\end{minipage}
\end{figure*}

\subsection{Model Analysis}
\label{subsec:analysis}

All experiments are conducted on the ScanObjectNN dataset in rotated views.
Please refer to the Appendix for detailed experimental analyses.

\noindent\textbf{PANet component ablations}
We evaluate the role of the cross-view association module and  $\mathcal{L}_{awe}$, as shown in Tab.~\ref{tab:component_ablation}.
The CVA module plays a positive role in enhancing the performance of PANet, improving instance accuracy and class accuracy by 0.69\% and 1.23\%.
The results show that the CVA module enables each view to more effectively integrate information from other views, further enhancing the representation capability of the view features.
Our method also improves by 2.40\% and 1.76\% in two metrics by introducing the $\mathcal{L}_{awe}$ constraint.
In addition, we also conduct correlation analysis between part features $\underline{\textbf{P}}$, as shown in Fig.~\ref{fig:vis_part}.
We observe that without applying $\mathcal{L}_{awe}$ constraints, the correlation between parts exhibits a flat similarity matrix, indicating that the parts are unvaried and cannot be distinguished in the feature space.
In contrast, the part features become more diverse with $\mathcal{L}_{awe}$ constraint.
Therefore, we conclude that $\mathcal{L}_{awe}$ can supervise the perception of parts to obtain diverse part features.

\begin{table*}[tp]
        \caption{Explore the impact of different viewpoint sampling on performance.}
        \scriptsize
        \centering
        \setlength{\tabcolsep}{1.6mm}{
        \begin{tabular}{cccccc}
        \toprule
        \multirow{2}{*}{Dataset} & \multirow{2}{*}{Method} & \multicolumn{2}{c}{Aligned}  & \multicolumn{2}{c}{Rotated}      
                \\ 
        \cmidrule{3-6}
        &  & {Class Acc.} & {Ins. Acc.}  & {Class Acc.} & {Ins. Acc.}    \\ \midrule
        \multirow{2}{*}{ModelNet40} & \textbf{random sampling}     &    96.37\%    &     97.53\%   &  90.69\%  &   92.23\% \\
        & \textbf{furthest point sampling}       &    95.44\%    &      97.41\%     &  89.73\%   &   92.26\% \\
        \midrule
        \multirow{2}{*}{ScanObjectNN} & \textbf{random sampling}      &    91.77\%    &     93.48\%    &  84.08\%  &   86.45\% \\
        & \textbf{furthest point sampling}       &    92.97\%    &     93.48\%    &  84.80\%  &   85.93\% \\
        \bottomrule
        \end{tabular}}
\label{tab:viewpint_sampling}
\end{table*}
\noindent\textbf{Influence of Different Viewpoint Sampling}
We introduce two different sampling methods: random sampling and furthest point sampling~(FPS).
Random sampling refers to the process of randomly selecting 10 to 20 views from each object.
This method is closer to real-world scenarios, and we showcased PANet's performance in Tab.~\ref{tab:scanobjectnn},~\ref{tab:align_modelnet},~\ref{tab:rotation_modelnet},~\ref{tab:rgbd} based on this sampling method.
FPS selects 10 to 20 views based on the furthest point sampling principle. 
It provides a uniform sampling method that covers every orientation of the 3D objects.
As shown in Tab.~\ref{tab:viewpint_sampling}, we observed that the performance of random sampling and FPS across different datasets is nearly similar.
This result shows that PANet is not sensitive to the choice of sampling methods.
The model demonstrates robustness in adapting to changes in viewpoint caused by different sampling methods while maintaining stable performance.

\begin{table}[t]
    \center
    \caption{Explore the impact of number of views and backbones on performance.}
    \scriptsize
    \setlength{\tabcolsep}{2.5mm}
    \resizebox{10.2cm}{!}{
        \begin{tabular}{c c c c c c}
        \toprule
        \multirow{2}{*}{\textbf{Backbone}}  & \multicolumn{5}{c}{\textbf{Views}} \\
        \cline{2-6}       & \multirow{1}[2]{*}{1 views} & \multirow{1}[2]{*}{5 views} & \multirow{1}[2]{*}{10 views} & \multirow{1}[2]{*}{15 views} & \multirow{1}[2]{*}{20 views} \\
        \midrule
        \multirow{1}{*}{\textbf{resnet18}}  &45.63\% & 70.84\% & 75.66\% & 79.90\% & \textbf{82.85\%} \\
        \midrule
        \multirow{1}{*}{\textbf{resnet50}}  & 67.07\% & 79.59\% & 82.50\% & 85.93\% & \textbf{86.62\%} \\
        \bottomrule 
        \end{tabular}%
    }
  \label{tab:view_number}
\end{table}
\noindent\textbf{Influence of the Number of the Views}
We further evaluate PANet by considering different backbone networks and numbers of views, as shown in Tab.~\ref{tab:view_number}. 
In previous studies, such as View-GCN~\cite{view-gcn}, because view aggregation leads to feature loss.
Choosing a stronger backbone network may not necessarily lead to performance improvement in certain dataset.
However, our observations reveal that the choice of backbone significantly affects PANet performance.
This is because PANet directly integrates multiple parts to characterize 3D objects. 
Therefore, a stronger backbone network is more advantageous for perceiving part features.
Furthermore, we notice a positive correlation between the number of views and the performance of PANet. 
It is worth noting that previous studies on multi-view aggregation methods, such as MHBN~\cite{MHBN}, MVLADN~\cite{MVLADN}, and GVCNN~\cite{gvcnn}, have indicated that increasing the number of views does not necessarily lead to better representation of 3D objects.
PANet combines multiple parts for representation, and the APR module can also filter out some redundant parts.
This means that PANet is not affected by redundant views.
Our experimental results also verify this idea.

\begin{table*}[tb]
\parbox[t]{.48\linewidth}{\centering
\caption{Exploring the effect of the number of attention maps, ${\rm M}$ denotes the number of attention maps per view.}
\setlength{\tabcolsep}{5pt}
\scriptsize
\begin{tabular}{c c c}
            \toprule
             \#Attention Maps           &  Class Acc.  & Ins. Acc.  \\
            \midrule
            ${\rm M = 16}$  & {84.57}\%         & {86.28}\%       \\ 
            ${\rm M = 32}$  & {85.36}\%         & {87.14}\%       \\ 
            ${\rm M = 64}$  & \textbf{{86.50}\%}         & \textbf{{87.31}\%}       \\ 
            ${\rm M = 128}$ & {86.38}\%         & {87.14}\%       \\ 
            \bottomrule
\end{tabular}
\label{tab:attention_map}
}
\hfill
\parbox[t]{.48\linewidth}{
\centering
\caption{Exploring the effect of the number of part tokens, $\rm L$ indicates the number of parts.}

\setlength{\tabcolsep}{5pt}
\scriptsize
\begin{tabular}{c c c}
            \toprule
             \#Number of Parts           &  Class Acc.  & Ins. Acc.  \\
            \midrule
            ${\rm L = 1}$  & {85.50}\%         & {86.28}\%       \\ 
            ${\rm L = 8}$  & {86.20}\%         & {87.14}\%       \\ 
            ${\rm L = 16}$ & \textbf{{86.50}\%}         & \textbf{{87.31}\%}       \\ 
            ${\rm L = 32}$ & {85.63}\%         & {86.31}\%       \\ 
            \bottomrule
\end{tabular}
\label{tab:parts}
}
\end{table*}

\noindent\textbf{Influence of the Number of the Attention Maps and Parts}
We show the impact of the number of attention maps $\rm M$ and the number of part tokens $\rm L$, as shown in Tab.~\ref{tab:attention_map} and Tab.~\ref{tab:parts}.
The number of attention maps  $\rm M$ determines the discriminative region in each view, and the part tokens $\rm L$ determine the number of global parts.
More  $\rm M$ and $\rm L$ usually contribute to better performance.
MACNN~\cite{macnn} and WSDAN~\cite{wsdan} report similar conclusions.
The experiments in Tab.~\ref{tab:attention_map} show that as the value of $\rm M$ increases, the classification accuracy gradually improves.
When $\rm M = 64$, the performance tends to stabilize.
The results in Tab.~\ref{tab:parts} show that the performance improves as the number of parts increases and reaches a peak at $\rm L=16$.
Compared with a single part, multiple parts provide richer information and enhance the generalization of the model.

\subsection{Visualization}
\label{subsec:vis}
\begin{figure*}[b]
    \centering   
    \includegraphics[width=\linewidth]{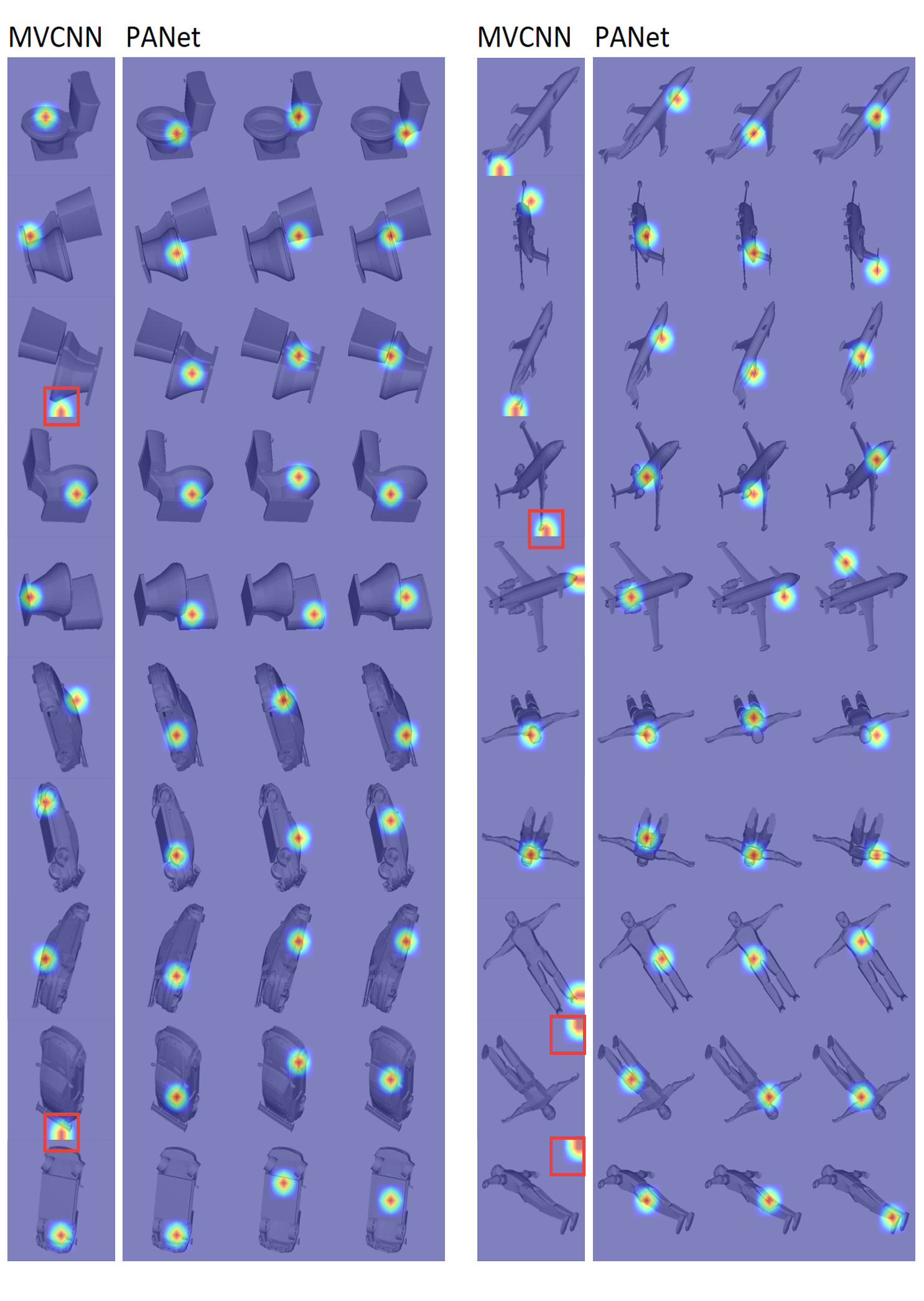}
     \caption{Comparing the semantic differences between view-based aggregation methods and PANet in feature extraction.}
    \label{fig:vis}
\end{figure*}

We randomly select some 3D models from a test set and compare the semantic differences between the features extracted by MVCNN and PANet, as shown in Fig.~\ref{fig:vis}.
MVCNN~\cite{mvcnn} is a classic feature aggregation method.
We notice that the features extracted by MVCNN are not entirely focused on the object.
This is because the uneven weight distribution leads to view aggregation methods ignoring certain view information.
It is worth noting that PANet does not discard any views but instead considers the features of each view, leveraging the valuable information within each view.
We can observe that different views emphasize various part locations, with each view capturing multiple discriminative regions (\eg~, the airplane nose, body, and wings).
This observation further validates the diversity of captured parts.
Furthermore, we observe that the view aggregation method does not explicitly model the unique attributes of 3D objects, leading to feature ambiguity.
In contrast, part-based representation reveals the inherent properties of objects, enhancing the interpretability of features and making recognition easier.
Please refer to the Appendix for more visualization results.

\section{Conclusion}
In this paper, we propose a novel part-aware network (PANet), which is a part-based representation.
Compared with view aggregation methods, this part-based representation has properties such as viewpoint invariance and rotation robustness, which can more effectively address the 3D object recognition problem under arbitrary views.
Moreover, we proposed the part-aware loss to ensure that part features can better focus on the object, thereby fully utilizing the information from each view.
We evaluate the performance of PANet on several 3D datasets, including ScanObjectNN, ModelNet, and RGBD.
The experimental results show that PANet has made significant progress, especially in arbitrary views.
We also conduct ablation and visualization experiments to study the impact of different components in our method.
It is worth noting that in these experiments, we found that our method is not affected by redundant views~(Sec.~\ref{subsec:analysis}), and the part features can better reveal the inherent properties of objects~(Sec.~\ref{subsec:vis}).
In conclusion, the experimental results prove the effectiveness of our method.\\

\noindent\textbf{Acknowledgement} 
\small
{This research is supported by the Shenzhen Science and Technology Plan Project No. KJZD20230923113800002 and Shenzhen LongHua Fundamental Research Program No. 10162A20230325B73A546.
}

\clearpage  

%
%
\bibliographystyle{splncs04}
\bibliography{main}
\end{document}